\documentclass[11pt]{article}  

\usepackage[utf8]{inputenc}
\usepackage[T1]{fontenc}
\usepackage{microtype}           
\usepackage{lmodern}             
\usepackage{amsmath, amssymb}    
\usepackage{physics}             
\usepackage{graphicx}            
\usepackage{booktabs}            
\usepackage{array}               
\usepackage{tabularx}            
\usepackage{multirow}            
\usepackage{adjustbox}           
\usepackage{float}               
\usepackage{tcolorbox}           
\tcbuselibrary{skins,breakable}
\usepackage{afterpage}           
\usepackage{url}                 
\usepackage{xurl}                
\newcolumntype{C}{>{\centering\arraybackslash}X}  
\newcolumntype{L}{>{\raggedright\arraybackslash}X} 
\newcolumntype{R}{>{\raggedleft\arraybackslash}X}  

\usepackage{tikz}
\usetikzlibrary{arrows.meta, positioning, matrix, calc, 
                shapes.geometric, decorations.pathreplacing}
\tikzset{>=Stealth}            

\usepackage{pifont}            
\newcommand{\cmark}{\checkmark}        
\newcommand{\xmark}{\ding{55}}        

\usepackage[top=1in, bottom=1in, left=0.75in, right=0.75in]{geometry}
\usepackage[font=small,labelfont=bf]{caption}    
\newcommand{\arxivnote}[1]{%
    \begin{tcolorbox}[
        colback=gray!5,
        colframe=gray!30,
        sharp corners,
        boxrule=0.5pt,
        left=4pt,
        right=4pt,
        top=2pt,
        bottom=2pt,
        fontupper=\small  
    ]
    \textbf{arXiv Note:} #1
    \end{tcolorbox}%
}

\usepackage[numbers,sort&compress]{natbib}
\usepackage[colorlinks=true,     
            linkcolor=blue,
            citecolor=blue,
            urlcolor=blue,
            pdfborder={0 0 0}]{hyperref}

\hypersetup{
    pdftitle={Quantum Abduction: A New Paradigm for Reasoning Under Uncertainty},
    pdfauthor={Remo Pareschi},
    pdfsubject={Artificial Intelligence, Quantum Cognition, Abductive Reasoning},
    pdfkeywords={quantum abduction; abductive reasoning; quantum cognition; semantic embeddings; forensic reasoning; scientific discovery},
    pdfcreator={LaTeX with hyperref package},
    pdfproducer={pdfLaTeX}
}

\begin{document}

\title{Quantum Abduction: A New Paradigm for Reasoning Under Uncertainty}

\author{Remo Pareschi\\
        \small Stake Lab, University of Molise\\
        \small 86100 Campobasso, Italy\\
        \small \href{mailto:remo.pareschi@unimol.it}{remo.pareschi@unimol.it}\\
        \small ORCID: 0000-0002-4912-582X
}

\date{}  

\maketitle

\arxivnote{This version corresponds to the final published article in 
\textit{Sci} 2025, Vol.~7, Issue~4, Article~182. 
DOI: \href{https://doi.org/10.3390/sci7040182}{10.3390/sci1010000}. 
Formatted for arXiv compliance by removing publisher-specific formatting while 
preserving all scholarly content, corrections, figures, and references.}

\begin{abstract}
\noindent
Abductive reasoning---the search for plausible explanations---has long been central to human inquiry, from forensics to medicine and scientific discovery. Yet formal approaches in AI have largely reduced abduction to eliminative search: hypotheses are treated as mutually exclusive, evaluated against consistency constraints or probability updates, and pruned until a single ``best'' explanation remains. This reductionist framing fails on two critical fronts. First, it overlooks how human reasoners naturally sustain multiple explanatory lines in suspension, navigate contradictions, and generate novel syntheses. Second, when applied to complex investigations in legal or scientific domains, it forces destructive competition between hypotheses that later prove compatible or even synergistic, as demonstrated by historical cases in physics, astronomy, and geology. This paper introduces \textbf{quantum abduction}, a non-classical paradigm that models hypotheses in superposition, allowing them to interfere constructively or destructively, and collapses only when coherence with evidence is reached. Grounded in quantum cognition and implemented with modern NLP embeddings and generative AI, the framework supports dynamic synthesis rather than premature elimination. For immediate decisions, it models expert cognitive processes; for extended investigations, it transforms competition into ``co-opetition'' where competing hypotheses strengthen each other. Case studies span historical mysteries (Ludwig II of Bavaria, the ``Monster of Florence''), literary demonstrations (\textit{Murder on the Orient Express}), medical diagnosis, and scientific theory change. Across these domains, quantum abduction proves more faithful to the constructive and multifaceted nature of human reasoning, while offering a pathway toward expressive and transparent AI reasoning systems.
\end{abstract}


\section{Introduction}

\subsection{Motivation} 
 
Many critical domains---criminal investigation, clinical diagnostics, scientific discovery\linebreak---require reasoning under ambiguity, contradiction, and evolving evidence. Classical abduction, as formalized in logic, Bayesian inference, or set-covering models, is typically reduced to eliminative selection: hypotheses compete within a fixed space until one prevails. This reductionist approach fails on two distinct timescales.

First, in immediate decision contexts (emergency medicine, real-time diagnostics), it does not reflect how human experts actually reason, often maintaining competing explanations in play until evidence forces convergence. Second, in extended investigative contexts (complex criminal cases, scientific theory development), the eliminative paradigm creates counterproductive competition among research teams, investigative units, or theoretical schools, when collaborative ``co-opetition''---where competing hypotheses inform and strengthen each other's development---would be more fruitful.

\subsection{Dual Role of Quantum Abduction}
Quantum abduction addresses both challenges through a unified framework. For immediate decisions, it models the parallel processing observed in expert cognition, where multiple explanatory threads remain active until collapse is necessary. For extended investigations, it transforms competitive hypothesis testing into a coordinated exploration where: 

\begin{itemize}
\item Different investigative teams pursue distinct but informationally entangled hypotheses
\item Evidence gathered for one hypothesis can constructively or destructively interfere with others
\item The framework acts as a meta-cognitive coordinator, tracking how partial insights from competing approaches might synthesize into richer explanations
\item Resources are allocated based on current amplitude distributions rather than premature winner-takes-all decisions
\end{itemize}

This ``co-opetition'' model is particularly valuable in complex criminal investigations (where tunnel vision on a single suspect can blind investigators to alternative leads) and scientific research (where premature paradigm commitment can delay breakthrough syntheses for decades). 

\subsection{Nature and Originality of This Work}
This paper presents a research framework with initial conceptual validation for a novel approach to abductive reasoning. The framework, termed quantum abduction, is entirely original in its specific formulation, though it builds upon established work in quantum cognition \cite{busemeyer2012quantum} and extends our prior research on entangled heuristics \cite{ghisellini2025entangled}. While the latter focuses on strategic concept creation through ontological synthesis, quantum abduction is epistemically grounded---it seeks the best explanation of a single determinate reality rather than creating new conceptual ontologies. The use of quantum formalism as a computational metaphor for cognitive processes, while not new, is applied here in an original way to the specific problem of abductive inference.

From a computational and formal standpoint, we adopt the Hilbert space formalism from quantum cognition to model
interference, contextuality, and order effects in explanation; see Section~\ref{sec:why_quantum} for the full rationale and limits.

\subsection{Structure of the Paper}
The remainder of the paper is organized as follows.
Section~\ref{sec:background} motivates the use of Hilbert space formalisms for abductive inference and clarifies the scope and limits of the quantum analogy.
Section~\ref{sec:premises} introduces the quantum abduction framework and its computational components.
Section~\ref{sec:casestudies} illustrates its use across forensic, clinical, literary, and scientific contexts.
Section~\ref{sec:benchmark_ludwig} provides an initial empirical assessment using a minimal benchmark.
Section~\ref{sec:related} contrasts the approach with logic-based, Bayesian, set-covering, and other alternative abductive traditions.
Section~\ref{sec:conclusion} concludes with implications, limitations, and directions for future development within the broader research program. A glossary of key terms is provided in Appendix~\ref{app:glossary}, and formal specifications are consolidated in Appendix~\ref{app:formal}.

\section{Background}
\label{sec:background}
\subsection{Classical Abduction}
Abductive reasoning---variously termed abduction, abductive inference, or retroduc-\linebreak tion---is a form of logical inference that proposes the most plausible explanation for a given set of observations. The concept was first articulated by the American philosopher and logician Charles Sanders Peirce \cite{peirce1878} in the late 19th century, who framed abduction alongside deduction and induction as one of the three fundamental modes of reasoning. More recent accounts emphasize its cognitive grounding and role as a general reasoning pattern, both in scientific discovery and in everyday inference \cite{Magnani09,reid2019}.

Unlike deduction, which guarantees the truth of its conclusions if the premises are true, abduction yields hypotheses that are tentative and provisional. Its conclusions are best understood as plausible guesses or informed conjectures, often described as “best available” or “most likely” explanations. Abduction also differs from induction: whereas induction generalizes from particular cases to broader regularities, abduction remains tied to specific observations, advancing hypotheses without claiming universality.

From the 1980s onward, the concept of abduction has garnered increasing interest in various applied fields, including law, clinical practice, computer science, and artificial intelligence. Diagnostic expert systems, for example, often used abduction to generate candidate explanations for observed symptoms or evidence. However, these implementations revealed both practical and theoretical limits. Practically, they were constrained by the computational resources of the time and by the lack of robust learning mechanisms. Theoretically, they were limited by the eliminative character of classical abduction itself: hypotheses are typically treated as mutually exclusive, and the reasoning process progresses by discarding inconsistent candidates until a single “best” explanation remains.

This eliminativist dynamic, grounded in the principles of non-contradiction and the excluded middle, creates brittleness precisely in situations where human experts exhibit flexibility---namely, in contexts of contradictory, incomplete, or entangled evidence. Classical abduction struggles to sustain multiple explanatory threads in suspension, forcing early closure at the expense of richer, hybrid accounts. These shortcomings provide the rationale for exploring quantum abduction, a framework that models inference in terms of superposition and interference, allowing explanatory alternatives to coexist and interact before eventual resolution.

\subsection{Quantum Cognition}
Quantum cognition is a theoretical framework that applies mathematical tools from quantum mechanics, not to claim that the brain is a quantum system, but to capture cognitive phenomena that resist classical probabilistic modeling. Introduced in the early 2000s \cite{busemeyer2011quantum,busemeyer2012quantum}, the approach has been used to explain puzzling features of human reasoning, such as order effects in decision-making, violations of classical probability axioms, and the coexistence of seemingly incompatible beliefs.

At its core, quantum cognition replaces classical probability spaces with Hilbert spaces, in which mental states are represented as superpositions of potential judgments or beliefs. Observations, questions, or contextual cues act as projection operators, modifying the cognitive state and leading to outcomes that display interference or entanglement effects. For instance, in survey responses, the probability of answering “yes” to two sequential questions often depends on their order---an effect naturally modeled as the collapse of a quantum superposition, but poorly explained by classical Bayesian frameworks.

We use these tools as a \emph{modeling formalism} for non-classical aspects of reasoning (interference, contextuality),
not as a claim about quantum hardware or ontological indeterminacy; detailed clarifications appear in Section \ref{sec:why_quantum}.


\subsection{Entangled Heuristics and Strategic Inference}
The work entitled ``Entangled Heuristics for Agent-Augmented Strategic Reasoning''~\cite{ghisellini2025entangled} 
 introduces a hybrid deductive reasoning architecture in which strategic heuristics are activated and composed through semantic entanglement. In that context, heuristics extracted from strategic texts (e.g., military treatises, corporate frameworks) are treated not as discrete rules but as semantically interdependent potentials, with transformer models employed to measure and exploit these semantic interdependencies \cite{ghisellini2024heuristics}. Conflicting or complementary heuristics are not eliminated but rather fused, leading to the emergence of novel and conceptually richer heuristic syntheses. This process embodies a quasi-ontological creativity: new strategic concepts arise through heuristic entanglement rather than being selected from a fixed menu of pre-existing options.

While the philosophical stance of that project was deliberately constructivist---treating inference as a form of creation---quantum abduction maintains a stronger epistemic orientation. Here, hypotheses represent real but uncertain explanatory elements about a determinate world. The entanglement mechanism remains valuable; however, embedding hypotheses in a Hilbert-style semantic space, modeling their interferences, and enabling synthetic recombination provide the computational machinery needed to capture inference dynamics in uncertain contexts.

Quantum abduction can thus be seen as a specialized adaptation of the entangled heuristics framework:

\begin{itemize}
\item It inherits the machinery of semantic superposition and interference.
\item But it applies them in service of \emph{explanatory convergence with reality}, rather than conceptual innovation.
\end{itemize}

This lineage clarifies how formal resources developed for strategic deduction can be repurposed to manage the epistemic complexities of abduction---enabling explanations to interplay, synthesize, or collapse only when evidence warrants it, rather than through premature elimination.

Moreover, the complementarity between two traditions is evident: quantum cognition models context effects, superposition, and interference in human judgment, while generative AI enables compositional and narrative synthesis. Their integration paves the way for quantum-abduction engines capable of generating natural-language explanations of complex entanglements. Such explanations allow ordinary human users to make sense of situations that would otherwise appear inextricably heterogeneous or \mbox{hopelessly contradictory}.

Figure~\ref{fig:pipeline_comparison} anticipates Section~\ref{sec:premises} by contrasting
eliminative abduction with superposition interference.

\afterpage{
\begin{figure}[htbp]
\centering
\resizebox{0.9\textwidth}{!}{
\begin{tikzpicture}[
  >=Stealth,
  node distance=7mm,
  box/.style={draw, rectangle, rounded corners=2pt, align=center, inner sep=4pt, text width=35mm, minimum height=10mm, font=\small},
  head/.style={font=\bfseries\small},
  shorten >=2pt, shorten <=2pt
]

\begin{scope}[xshift=0cm]
\node[head] (ctitle) at (0,3.8) {Classical Abduction Pipeline};
\node[box, minimum width=35mm, fill=blue!10] (c1) at (0,3.0) {Input: Hypotheses + Observations};
\node[box, fill=purple!10] (c2) [below=of c1] {Encoding: Logical or Symbolic Representation};
\node[box, fill=purple!10] (c3) [below=of c2] {Evaluation: Consistency Checks};
\node[box, fill=purple!10] (c4) [below=of c3] {Elimination of Inconsistent Hypotheses};
\node[box, fill=yellow!20] (c5) [below=of c4] {Output: Best-Supported Hypothesis $H^{*}$};
\draw[->] (c1) -- (c2);
\draw[->] (c2) -- (c3);
\draw[->] (c3) -- (c4);
\draw[->] (c4) -- (c5);
\end{scope}

\begin{scope}[xshift=8.5cm]
\node[head] (qtitle) at (0,3.8) {Quantum Abduction Pipeline};
\node[box, minimum width=35mm, fill=blue!10] (q1) at (0,3.0) {Input: Hypotheses + Observations};
\node[box, fill=purple!10] (q2) [below=of q1] {Semantic Embedding\\(Hilbert space Vectors)};
\node[box, fill=purple!10] (q3) [below=of q2] {Projection: Observations $\mapsto$ Hypotheses};
\node[box, fill=purple!10] (q4) [below=of q3] {Interference Matrix: Constructive / Destructive};
\node[box, fill=purple!10] (q5) [below=of q4] {Amplitude Dynamics: Continuous Re-weighting};
\node[box, fill=yellow!20] (q6) [below=of q5] {Output: Collapse to Dominant or Hybrid Explanation};
\draw[->] (q1) -- (q2);
\draw[->] (q2) -- (q3);
\draw[->] (q3) -- (q4);
\draw[->] (q4) -- (q5);
\draw[->] (q5) -- (q6);
\end{scope}

\end{tikzpicture}}
\caption{Computational pipelines: Classical abduction 
 prunes the hypothesis space through elimination. Quantum abduction embeds hypotheses in a semantic Hilbert space, models interference, and converges via amplitude dynamics to a dominant or hybrid synthesis.}
\label{fig:pipeline_comparison}
\end{figure}}

\section{Computational Premises for Quantum Abduction}
\label{sec:premises}

Classical abduction, as outlined in Section~\ref{sec:background}, typically assumes a fixed hypothesis set and an eliminative search process: contradictions are resolved by discarding candidates until a single explanation remains. Rather than reiterating these limitations, we now turn to the computational consequences of moving beyond them.

Quantum abduction shares the classical commitment to a determinate underlying reality---only one sequence of events has actually occurred---but departs from eliminativism in its representation of uncertainty. Hypotheses are maintained in an \emph{epistemic superposition}, with amplitudes that evolve as evidence accumulates. Observations act as projections in a semantic Hilbert space; interactions among hypotheses are modeled through an interference matrix that captures reinforcement, contradiction, or semantic overlap. Collapse marks explanatory convergence, which may result in a dominant or hybrid explanation rather than a single survivor from a fixed menu.

This section introduces the operational machinery that enables these dynamics, centered on the mathematical summary in Box~1. We treat Hilbert space tools---vectors, projections, interference---not as ontological commitments but as a computational formalism that resists premature elimination and preserves information until evidence warrants decisive convergence. Full derivations appear in Appendix~\ref{app:formal}.

\subsection{Why Quantum?---Interpretational Clarifications} 

\label{sec:why_quantum}

As anticipated in the Introduction, “quantum” is used here in an \emph{epistemic and computational} sense rather than an ontological one. Section~\ref{sec:background} gave the conceptual motivation; here we state the implications at a more formal level and show how they guide the \mbox{computational design}.

\paragraph{Three Distinct Notions} 

\begin{itemize}
\item \textbf{Quantum cognition}: Hilbert space structure to model belief states and their context-sensitive evolution (this paper). 
\begin{itemize}
    \item \emph{Intuition:} A Hilbert space offers a geometric setting in which each hypothesis is a vector and each observation acts as a projection. Inner products encode explanatory alignment; interference terms capture how alternatives can reinforce or attenuate one another. This yields contextuality and order effects providing the non-classical compositionality needed for abductive reasoning \mbox{under contradiction}.
  \end{itemize}
\item \textbf{Quantum computation}: algorithms on quantum hardware (not used here).
\item \textbf{Classical simulation of quantum-like dynamics}: classical hardware implementing interference/superposition mathematics (our regime).
\end{itemize}

\paragraph{Why the Formalism Matters Here}
The formalism captures three features of explanatory reasoning that classical abduction struggles with:
\begin{enumerate}
\item \emph{Interference} of explanatory alternatives, yielding non-classical updates (violations of the law of total probability when hypotheses interact);
\item \emph{Contextuality}, where the plausibility of a hypothesis depends on which other hypotheses are co-considered;
\item \emph{Order effects} in evidence evaluation, as documented in cognitive experiments \cite{busemeyer2011quantum}.
\end{enumerate}

Superposition and collapse are \emph{epistemic}: coexistence of competing explanations prior to convergence; collapse as emergence of a coherent synthesis under sufficient constraints.

\paragraph{Relation to Entangled Heuristics}
Methodologically, we share machinery with \textit{Entangled Heuristics for Agent-Augmented Strategic Reasoning} \cite{ghisellini2025entangled} (transformers encode semantic interdependence; interference drives composition). The interpretational aim differs:
\begin{itemize}
\item In \cite{ghisellini2025entangled}, semantic superposition supports \emph{ontological innovation} (new strategic constructs).
\item Here, superposition tracks \emph{epistemic uncertainty about a determinate reality}.
\end{itemize}

Thus, the same tools (semantic embeddings, transformer-mediated similarity, interference dynamics) serve ontological synthesis there, and epistemic inference here.

\begin{tcolorbox}[title=Quantum 
 Abduction: Core Math Capsule, colback=gray!2, colframe=black!25, breakable]
\small
\textbf{Abductive state}:
$\ket{\Psi^{(t)}}=\sum_i \alpha_i^{(t)} \ket{H_i}$,\quad $\sum_i |\alpha_i^{(t)}|^2=1$.

\textbf{Projection (evidence)}:
$e_i^{(t)}=\langle H_i \mid O_j\rangle=\cos(\mathbf{h}_i,\mathbf{o}_j)$.

\textbf{Interference}:
$I_{ij}=\kappa_{ij}\cdot \text{sim}(H_i,H_j)$,\; $\kappa_{ij}\in[-1,1]$.

\textbf{Update + normalize}:
$\tilde{\alpha}_i^{(t+1)}=\alpha_i^{(t)}+\eta\!\left(e_i^{(t)}+\sum_{k\neq i} I_{ik}\alpha_k^{(t)}\right)$,\;
$\alpha^{(t+1)}=\tilde{\alpha}^{(t+1)}/\|\tilde{\alpha}^{(t+1)}\|_2$.

\textbf{Collapse} when $C(\Psi)=\max_i |\alpha_i|^2>\tau$ (single or hybrid via mix).
Formal details: Appendix~\ref{app:formal}, Equations~(\ref{eqA:state})–(\ref{eqA:collapse}).
\end{tcolorbox}

\paragraph{Transformers and LLMs (Conceptual Primer)}
Hypotheses and observations are expressed in natural language; we therefore need a representation that preserves semantic structure. Transformer encoders map text to high-dimensional vectors, where semantic proximity corresponds to geometric proximity. This does not entail that transformers “understand” the hypotheses; rather, they supply a \emph{coordinate system} in which superposition and interference can be computed. LLMs, in turn, \emph{articulate} hybrid explanations in natural language; they do not decide scores.

\subsection{Transformer and LLM Integration as Epistemic Infrastructure}
\label{sec:transformers_llms}

\paragraph{Embedding Model}
We use the transformer model Sentence-BERT \cite{reimers2019sentencebert}
(e.g., all-MiniLM-L6-v2, $d{=}384$) to encode hypotheses $H_i$ and
observations $O_j$ into a shared semantic space:
$\mathbf{h}_i = \mathrm{SBERT}(\text{text}(H_i))$ and
$\mathbf{o}_j = \mathrm{SBERT}(\text{text}(O_j))$.
Results are largely invariant to the specific SBERT variant up to rotation;
higher-dimensional models increase fidelity at modest cost.

\paragraph{Clarifying the Objects Involved}
Each hypothesis $H_i$ and observation $O_j$ begins as a text string.
These are not numerical objects. To use them computationally, we convert them
into fixed-length numerical vectors using the SBERT encoder.

\paragraph{What SBERT Does (Intuitive Explanation)}
SBERT is a transformer-based model that maps any piece of text---ranging from
a single word to a short paragraph---into a vector in $\mathbb{R}^{384}$.
All embeddings have the same dimensionality regardless of the input length.
This uniformity is why cosine similarity between
$\mathbf{h}_i$ and $\mathbf{o}_j$ is well-defined.

In principle, long hypotheses could be decomposed into clause- or
sentence-level embeddings and then pooled (e.g., by weighted averaging or
attention). For this initial framework we adopt the simpler and widely
used “one SBERT vector per text” strategy, leaving fine-grained composition
to future work.

\paragraph{Meaning of $H_i$, $\mathbf{h}_i$, and $e_i$}
To avoid ambiguity:
\begin{itemize}
\item $H_i$ = hypothesis as a linguistic description;
\item $\mathbf{h}_i$ 
 = its embedding in $\mathbb{R}^{384}$;
\item $O_j$ = observation as text;
\item $\mathbf{o}_j$ = its embedding in $\mathbb{R}^{384}$.
\end{itemize}

The projection score for hypothesis $i$ at time $t$ is
\[
e_i^{(t)} = \cos(\mathbf{h}_i, \mathbf{o}^{(t)}).
\]
It 
 has a single index because it measures the activation of one hypothesis,
even though its computation involves two embeddings.

\paragraph{Qualitative vs. Quantitative Projection Matrices}

The tables in Section~\ref{sec:casestudies} use qualitative marks
(v = supports, x = contradicts) purely for readability.  
In the actual computation, each entry is replaced by a cosine similarity
between SBERT embeddings, optionally modulated by an interference coefficient
$\kappa_{ij}$ \mbox{described below}.

\paragraph{Projection (Evidence Activation)}
Similarity $e_i^{(t)} = \cos(\mathbf{h}_i, \mathbf{o}_j)$ acts as a projection
operator, activating or suppressing amplitudes without forcing single-winner
selection.

\paragraph{Interference Estimation}
We set $I_{ij} = \kappa_{ij}\cdot \text{sim}(H_i,H_j)$ with cosine similarity over SBERT vectors. $\kappa_{ij}\in[-1,1]$ encodes exclusivity/complementarity and can be initialized (i) heuristically from domain priors, (ii) via weak supervision (optimize ECS/HQI on solved cases), or (iii) by expert elicitation smoothed through metric learning.

\paragraph{Amplitude Update (Single Step)}
\begin{verbatim}
Input: alpha[1..n], I[1..n,1..n], evidence e[1..n], eta in (0,1]
for i in 1..n:
    alpha_tilde[i] = alpha[i] + eta * ( e[i] + sum_{k != i} I[i,k] 
    * alpha[k] )
Normalize: alpha = alpha_tilde / ||alpha_tilde||_2
Collapse if max_i |alpha[i]|^2 > tau; else continue
\end{verbatim}

\paragraph{Mix Operator and LLM Articulation}
When multiple hypotheses retain high amplitudes and exhibit constructive interference, the system prepares a \emph{candidate hybrid} by combining their embeddings via a weighted semantic blend. The precise operator, its justification, and its role in case studies are elaborated in Section \ref{sec:mix}; at this point we note only that the LLM’s role is strictly limited to articulating the hybrid in natural language, not to scoring or selecting it.

\paragraph{Human-in-the-Loop}
Experts adjust $\kappa_{ij}$ (conflict/compatibility), calibrate collapse thresholds $\tau$, and edit LLM phrasing for clarity and faithfulness. This design follows a \emph{human-in-the-loop}  principle~\cite{BorgBottPare2025}, in which domain experts operate at a semantic rather than mathematical level. They do not need Hilbert space or amplitude calculus; their inputs are judgments about \emph{complementarity}, \emph{conflict}, or \emph{partial overlap} among hypotheses (i.e., signs and strengths for $\kappa_{ij}$, and acceptable collapse thresholds $\tau$). Within a broader \emph{centaurian} paradigm~\cite{centaurian2024}, the assistant provides formal control and interpretability over superposition/interference dynamics while preserving human agency over meaning and decision.

Full definitions, equations, and derivations are provided in Appendix~\ref{app:formal}; here we retain only the operational summary.

\subsection{The Mix Operator for Hypothesis Synthesis}
\label{sec:mix}

When several hypotheses retain high amplitudes and exhibit constructive interference, the system applies the \text{mix} operator to generate hybrid explanations: 

\begin{equation}
\text{mix}(H_i, H_j) = \lambda_{ij} \mathbf{h}_i + (1 - \lambda_{ij}) \mathbf{h}_j,
\qquad
\lambda_{ij} = \frac{|\alpha_i|^2}{|\alpha_i|^2 + |\alpha_j|^2}.
\end{equation}
The  natural-language articulation follows the LLM strategy above. Formal variants and derivations appear in Appendix~\ref{app:formal}.

\paragraph{Why This Matters in Practice}
\emph{Retention of incompatible evidence} (no premature pruning), \emph{dynamic reweighting} as data arrives, and \emph{hybrid resolutions} when interference is constructive are the core \mbox{computational benefits}.

\subsection*{How the Machinery Appears in Our Cases (Preview of Section \ref{sec:casestudies})}

\vspace{-7pt}
\begin{table}[H]
\caption{Preview of case-study mechanisms}
\label{tab:case-mechanisms}
\centering
\small
\setlength{\tabcolsep}{4pt}
\renewcommand{\arraystretch}{1.1}
\begin{tabular}{p{4.2cm}p{8.3cm}}
\toprule
\textbf{Case} & \textbf{Mechanism in Play} \\
\midrule
Ludwig II & 
$e_i$ from letters/reports boosts H1/H2/H3; positive $I_{12}$ and $I_{23}$ 
sustain a hybrid ``entangled conflict'' before collapse. \\
\addlinespace[0.2em]

Mostro di Firenze & 
Stable $e$ for weapon/M.O.\ raises H1/H3; vehicle variance drives $I_{12}>0$ 
(imitators); new DNA toggles $I_{35}$. \\
\addlinespace[0.2em]

Bossetti--Gambirasio & 
Strong $e$ for nuclear DNA on H1 opposed by mtDNA/Y-line (negative $I$ toward H1); 
superposition resists forced collapse. \\
\addlinespace[0.2em]

Botulism vs.\ GBS/MFS & 
Parallel high $e$ on H1/H2; defer collapse and allow treatment-on-superposition 
until decisive labs arrive. \\
\addlinespace[0.2em]

Drift $\rightarrow$ Tectonics & 
Long-lived superposition H1/H2; instruments increase $e$ on H1 and flip $I$ 
as mechanism emerges, prompting synthesis. \\
\bottomrule
\end{tabular}
\end{table}

\paragraph{Implementation Note}
Off-the-shelf SBERT encoders provide embeddings; $I_{ij}$ is initialized from similarity and tuned by priors or weak supervision; updates follow the capsule above. A short reference sketch and complexity notes appear in Appendix~\ref{app:formal}.

\subsection{Temporal Scales and Co-Opetition in Quantum Abduction}
\label{sec:coopetition}

\paragraph{Immediate Decision Contexts}
In time-constrained scenarios (such as emergency medicine, crisis response, and real-time diagnostics), experts naturally maintain multiple hypotheses in superposition, allowing for parallel consideration until evidence or urgency forces a collapse. Here, quantum abduction serves both as a \emph{descriptive model} of expert cognition and a \emph{prescriptive tool} for decision support.

\paragraph{Extended Investigative Contexts}
In long-duration investigations (complex criminal cases; scientific paradigm shifts), classical eliminativism fosters destructive competition. Quantum abduction coordinates “co-opetition”:
\begin{itemize}
\item \textbf{Hypothesis Allocation}: distribute resources proportionally to $|\alpha_i|^2$;
\item \textbf{Information Sharing}: evaluate evidence under $H_i$ for projection onto all active hypotheses;
\item \textbf{Interference Tracking}: monitor cross-effects via $I_{ij}$ and amplitude sensitivities;
\item \textbf{Synthesis Triggers}: detect constructive interference suggesting breakthrough hybrids.
\end{itemize}

\paragraph{Organizational Implementation}
\textls[-13]{Possible forms include investigation coordination centers, research consortia with shared amplitude tracking, and multi-unit intelligence analysis with \mbox{interference monitoring}}.

\paragraph{From Competition to Co-opetition}
The same mathematics ($\alpha$, $I$, $\tau$) scales to organizational coordination. A simple value decomposition is
\begin{equation}
\text{Value}_{\text{co-op}} = \sum_i |\alpha_i|^2 V_i + \sum_{i \neq j} I_{ij} \alpha_i \alpha_j S_{ij},
\end{equation}
where $V_i$ is the standalone value of $H_i$ and $S_{ij}$ is  
the synthetic value from interaction. The second term---absent in classical abduction---captures the added value of maintaining superposition.

\section{Case Studies}
\label{sec:casestudies}
We present case studies drawn from criminal investigation, clinical diagnostics, and scientific inquiry to illustrate how quantum abduction operates in the face of ambiguity, contradiction, and evolving evidence. Together, they span the full temporal range of the framework’s application: from real-time decision settings---such as emergency medical management where parallel hypotheses must remain active until action is unavoidable---to long-horizon contexts like scientific research programs or decades-long criminal investigations. Across these domains, the case studies show how quantum abduction sustains multiple explanatory lines in a structured superposition, countering the eliminative bias of classical approaches, and how it enables synthesis or collapse precisely when the evidence warrants it.

\subsection{Forensic Reasoning}
\label{sec:ludwig}
\subsubsection{Ludwig 
 II of Bavaria}

The mysterious death of King Ludwig II of Bavaria in 1886, together with his physician Dr.\ Gudden, is a paradigmatic case where classical abduction struggles and quantum abduction offers a richer interpretive space. The historical record presents conflicting evidence: eyewitness accounts of the bodies in shallow water, reports of blunt trauma on Gudden, Ludwig’s documented political deposition and alleged insanity, and the absence of a thorough autopsy. Classically, these contradictions enforce a premature binary: \emph{suicide} versus \emph{murder}.

Classical framing forces “suicide vs.\ murder”; the superpositional lens lets both narratives coexist until constraints compel synthesis. Concretely, political orchestration and personal despair can be treated as entangled contributors rather than exclusives, so that evidence about Gudden’s trauma, deposition dynamics, and Ludwig’s letters interact before any collapse. In our setting, this interaction yields an emergent $H_5$ (entangled conflict) that is not posited a priori but is constructed by interference among $H_1$--$H_3$; Section~\ref{sec:benchmark_ludwig} demonstrates this effect quantitatively using randomized evidence orderings.

Furthermore, at the \emph{hypothesis level}, quantum abduction supports hybrid explanations such as the following:

\begin{itemize}
    \item $H_1$: Suicide. Explains resignation but ignores Gudden’s injuries and political orchestration.
    \item $H_2$: Murder. Explains political context and Gudden’s trauma but leaves Ludwig’s fatalism underexplored.
    \item $H_3$: Struggle. Accounts for joint death and injuries but underplays political necessity.
    \item $H_4$: Medical accident. Considers sudden seizure or collapse but neglects political-military involvement.
    \item $H_5$: \textbf{Entangled conflict} (emergent from interference). This hypothesis is not defined a priori but emerges through the quantum abductive process from constructive interference between $H_1$, $H_2$, and $H_3$. It represents Ludwig resisting removal in a context where suicidal impulses, physical resistance, and political suppression became entangled, projecting onto most observations with high combined explanatory power.
\end{itemize}

Historical documents support this reframing, as illustrated in Table~\ref{tab:ludwig}. Ludwig’s letters to Richard Wagner and confidants reveal fatalistic tones and longing for escape, aligning with $H_1$; political records demonstrate ministerial urgency to depose him, aligning with $H_2$; Gudden’s ambiguous role aligns with $H_3$; reports of seizures align with $H_4$. Only $H_5$---the entangled conflict hypothesis---integrates them without contradiction. 

\begin{table}[H]
\caption{Hypothesis 
 projection matrix for Ludwig II case. Observations: O$_1$ = both died together, \mbox{O$_2$ = drowning}, O$_3$ = Ludwig did not drown, O$_4$ = Gudden trauma, O$_5$ = seizure, O$_6$ = dubious insanity, O$_7$ = political pressure. (\cmark = supports hypothesis; \xmark = contradicts hypothesis.)\label{tab:ludwig}}
\begin{tabularx}{\textwidth}{lCCCCCCC}
\toprule
\textbf{Hypothesis} & \textbf{O\textsubscript{1}} & \textbf{O\textsubscript{2}} & \textbf{O\textsubscript{3}} & \textbf{O\textsubscript{4}} & \textbf{O\textsubscript{5}} & \textbf{O\textsubscript{6}} & \textbf{O\textsubscript{7}} \\
\midrule
$H_1$ Suicide            & \cmark & \cmark & \xmark & \xmark & \xmark & \xmark & \xmark \\
$H_2$ Murder             & \cmark & \xmark & \cmark & \cmark & \xmark & \cmark & \cmark \\
$H_3$ Struggle           & \cmark & \xmark & \cmark & \cmark & \cmark & \xmark & \xmark \\
$H_4$ Medical            & \cmark & \xmark & \xmark & \xmark & \cmark & \xmark & \xmark \\
$H_5$ Entangled conflict & \cmark & (\cmark/\xmark) & \cmark & \cmark & \cmark & \cmark & \cmark \\
\bottomrule
\end{tabularx}
\end{table}

\vspace{-6pt}

Through quantum abduction, the Ludwig II case is reframed not as “who killed whom,” but as an entangled historical event in which political orchestration, personal despair, and interpersonal struggle jointly collapsed into the observed outcome. This illustrates the unique explanatory gain of a superpositional abductive framework over classical elimination.

\subsubsection{The “Monster of Florence”}

The “Mostro di Firenze” killings (1968--1985) are one of the most studied yet unresolved serial crime cases in Europe, involving eight double homicides of couples near Florence. Despite decades of investigation, no definitive suspect has been convicted. This case highlights the limitations of classical abduction and the interpretive power of a quantum abductive framework.

\paragraph{Observations}
Eight double homicides (1968–1985) with a consistent weapon/M.O.\ and post-mortem mutilations; heterogeneous vehicles reported near scenes; 2024 DNA update introduces an unknown external profile.

\paragraph{Competing Hypotheses}
$H_1$ single serial killer;\; $H_2$ multiple actors/imitators;\; $H_3$ transient actor with wide mobility;\; $H_4$ ritual/cult network;\; $H_5$ external predator.

\paragraph{Quantum Reframing}
Forensic consistency ($O_2$) boosts $H_1/H_3$; vehicular diversity activates $H_2$; the 2024 DNA boosts $H_5$. Constructive interference yields hybrids:
$H_{1,2}$ (core killer + imitators),
$H_{3,4}$ (transient acts appropriated into cultic narratives),
$H_{3,5}$ (external predator intersecting local patterns).

\paragraph{Quantum Advantage}
(i) No premature elimination of inconsistent cues;
(ii) contradictions become interferential structure, not noise;
(iii) composite explanations are first-class outcomes rather \mbox{than afterthoughts}.

\paragraph{Reframed Question}
The quantum abductive stance shifts from “Which hypothesis explains all?” to the following: 
\begin{quote}
    \emph{“Which entangled combination of agency, imitation, mobility, and symbolic behavior coherently explains the mixture of forensic consistency and observational variability in the Florence homicides?”}
\end{quote}

\paragraph{Illustrative Synthesis}
A plausible abductive synthesis might state the following:
\begin{quote}
    \emph{A 
 dominant transient actor with access to varied vehicles carried out the core attacks, whose stylistic and symbolic violence later inspired local imitators and speculative cultic interpretations. The consistency of weapon and mutilations suggests a single origin, while the observed vehicular variance and the emergence of new DNA indicate auxiliary actors or replication effects. The “Mostro” phenomenon thus emerges less as a single killer and more as an entangled field of violence, imitation, and unresolved identities.}
\end{quote}

\paragraph{Quantum Advantage}
Here, quantum abduction achieves the following:
\begin{itemize}
    \item Avoids premature exclusion of inconsistent evidence.
    \item Integrates contradictions as interference effects rather than eliminations.
    \item Supports composite explanations (killer + imitators + external unknown).
\end{itemize}

Thus, the framework captures the narrative multiplicity of the case, yielding explanatory depth beyond the eliminativist paradigm.

\subsubsection{Bossetti and the Yara Gambirasio Case}

The murder of Yara Gambirasio, a 13-year-old girl found dead in 2011 near Brembate di Sopra in the province of Bergamo, remains one of Italy's most controversial criminal cases. Massimo Giuseppe Bossetti was convicted in 2016 based on DNA evidence; this outcome of the criminal proceedings remains legally solid, despite numerous anomalies in the forensic chain and contradictory interpretations that continue to raise doubts about both the reliability of the evidence and the coherence of the investigative reasoning. In this case, a typical reductionist shift toward classical \emph{eliminativist} abduction---driven by the judicial necessity of closure---was required to uphold Bossetti's conviction. Yet, as with any conviction based purely on circumstantial or interpretative evidence, shadows remain. A quantum abductive approach, instead of enforcing premature closure, would keep these shadows in play, offering a fuller picture of the inferential landscape.

While the Bossetti case itself was not formally described in abductive terms within judicial proceedings, its evidential dynamics---mutually conflicting hypotheses, partial support, and iterative narrative reconstruction---are archetypally abductive. In this sense, our analysis is not a historical reconstruction but an abductive re-framing aimed at revealing how eliminative reasoning constrained the court’s interpretive space.

\paragraph{Observational Contradictions}

The prosecution rested on the matching of Bossetti’s nuclear DNA to the so-called \textit{Ignoto 1} profile recovered from the victim’s clothing. However, several anomalies emerged:

\begin{itemize}
  \item The \textbf{Y-chromosome haplotype} of Ignoto 1 did not match that of Bossetti’s known \mbox{male relatives}.
  \item \textls[-25]{The \textbf{mitochondrial DNA}  found on Yara’s clothes did not match Bossetti’s maternal line.}
  \item The amount of \textbf{nuclear DNA} was unusually high, despite its well-known tendency to degrade rapidly in environmental exposure.
  \item Conversely, \textbf{mitochondrial DNA}, typically far more stable and abundant in degraded samples, was found only in trace quantities.
\end{itemize}

This paradox constitutes an inversion of forensic expectations: nuclear DNA, normally fragile, persisted strongly, while mitochondrial DNA, normally resilient, was nearly absent. Classical reasoning treats such anomalies as marginal noise; a quantum abductive model instead elevates them to central dimensions of explanatory tension.

\paragraph{Limits of Classical Abduction}

Classical abductive reasoning proceeds as follows:
\begin{enumerate}
  \item Observation: Bossetti's nuclear DNA matches the crime scene sample.
  \item Inference Rule: If someone’s DNA matches a crime scene sample, then they were likely present.
  \item Abductive Conclusion: Bossetti was at the crime scene and likely committed \mbox{the murder}.
\end{enumerate}

This model cannot accommodate the contradictions above. To preserve logical consistency, it must eliminate or downplay conflicting evidence---mitochondrial mismatch, degradation anomalies, or kinship inconsistencies---resulting in a form of \emph{explanatory closure} that obscures deeper explanatory potentials.

\paragraph{Quantum Abductive Reframing}

Quantum abduction offers a framework in which conflicting observations need not be collapsed or discarded. Instead, each hypothesis is represented as a semantic vector in a conceptual Hilbert space, and each observation as a projection operator that activates or interferes with subsets of these hypotheses.

The hypothesis space may include the following:
\begin{itemize}
  \item $H_1$: Bossetti is the biological donor and the murderer.
  \item $H_2$: The donor is a relative or unregistered sibling.
  \item $H_3$: The DNA match results from laboratory contamination or error.
  \item $H_4$: Bossetti's DNA was planted or transferred indirectly.
  \item \textls[-25]{$H_5$: Degradation or environmental effects inverted nuclear vs.\ mitochondrial persistence.}
  \item $H_6$: Amplification or measurement bias distorted the detection process.
\end{itemize}

These hypotheses form an entangled field rather than discrete alternatives. For instance, $H_1$ and $H_2$ share semantic overlap (nuclear match) but diverge on kinship assumptions; $H_1$ is weakened by mitochondrial mismatch, which instead reinforces $H_2$ or $H_4$. The final explanatory synthesis emerges not from exclusion but from \mbox{interference-informed composition}.

\paragraph{Projection Matrix}

To visualize this entangled reasoning, in Table~\ref{tab:bossetti} we map hypotheses ($H_i$) against key observations ($O_j$):

\begin{table}[H]
\centering
\caption{Hypothesis projection matrix for the Bossetti--Gambirasio case. 
Observations: O$_1$ = nuclear DNA match, O$_2$ = Y-haplotype mismatch, 
O$_3$ = mitochondrial mismatch, O$_4$ = nuclear abundance anomaly, 
O$_5$ = mitochondrial scarcity, O$_6$ = contamination possibility, 
O$_7$ = court ruling affirming nuclear DNA integrity. ($\checkmark$ = supports hypothesis; $\times$ = contradicts hypothesis.)}
\label{tab:bossetti} 
\begin{tabular}{lccccccc}
\toprule
\textbf{Hypothesis} & \textbf{O\textsubscript{1}} & \textbf{O\textsubscript{2}} & \textbf{O\textsubscript{3}} & \textbf{O\textsubscript{4}} & \textbf{O\textsubscript{5}} & \textbf{O\textsubscript{6}} & \textbf{O\textsubscript{7}} \\
\midrule
$H_1$ Bossetti guilty      & $\checkmark$ & $\times$ & $\times$ & $\times$ & $\times$ & $\times$ & $\checkmark$ \\
$H_2$ Relative involvement & $\checkmark$ & $\checkmark$ & $\checkmark$ & $\times$ & $\times$ & $\times$ & $\times$ \\
$H_3$ Contamination        & $\checkmark$ & $\times$ & $\times$ & $\times$ & $\times$ & $\checkmark$ & $\times$ \\
$H_4$ Secondary transfer   & $\checkmark$ & $\times$ & $\checkmark$ & $\times$ & $\checkmark$ & $\times$ & $\times$ \\
$H_5$ Degradation anomaly  & $\times$ & $\times$ & $\checkmark$ & $\checkmark$ & $\checkmark$ & $\times$ & $\times$ \\
$H_6$ Measurement bias     & $\checkmark$ & $\times$ & $\checkmark$ & $\checkmark$ & $\checkmark$ & $\checkmark$ & $\times$ \\
\bottomrule
\end{tabular}
\end{table}

This matrix shows that no single hypothesis aligns with all observations. Instead, they form overlapping projections, sustaining multiple explanatory amplitudes.

\paragraph{Forensic Reinforcement and Recalibration}
On appeal, the Court of Cassation reaffirmed the integrity of nuclear DNA, which, in our model, acts as a strong projection, increasing $|\alpha_{H_1}|^2$, while applying destructive pressure to alternatives ($H_2$, $H_3$). As Figure~\ref{fig:bossetti_interference_split} shows, orthogonal constraints (mitochondrial and Y-line mismatches) prevent full collapse, explicitly separating \emph{judicial closure} from \emph{epistemic closure}.

Thus, a quantum abductive assistant might generate the following:

\begin{quote}
\emph{The 
 nuclear DNA match strongly implicates Bossetti or a close relative, but mitochondrial and Y-chromosome mismatches suggest genealogical or methodological gaps. The unusual abundance of nuclear markers, compared to the scarce mitochondrial evidence, indicates either secondary transfer, selective preservation, or amplification bias. While the Cassation ruling reinforces Bossetti’s responsibility, quantum abduction models this as a high-amplitude projection that suppresses but does not eliminate alternative hypotheses. The explanatory field remains entangled: identity, mechanism, and chain-of-custody processes co-determine the inferential landscape.}
\end{quote}

This reveals the epistemic innovation of quantum abduction: it accounts not only for forensic contradictions but also for the institutional dynamics of closure, showing how alternative amplitudes persist beyond verdicts. In this way, quantum abduction spans both modern forensic science and historical mysteries (as in the Ludwig II case), demonstrating its scope across contexts.

\begin{figure}[H]

\resizebox{\textwidth}{!}{%
\begin{tikzpicture}[
  >=Stealth,
  hyp/.style={draw, rectangle, rounded corners=2pt, align=center, inner sep=3pt,
              text width=3.8cm, minimum height=1.10cm, font=\normalsize},
  obs/.style={draw, rectangle, rounded corners=2pt, align=center, inner sep=3pt,
              text width=4.2cm, minimum height=1.0cm, font=\normalsize, fill=gray!10},
  act/.style={-Stealth, line width=1.2pt, black},        
  actstrong/.style={-Stealth, line width=1.8pt, black},  
  neg/.style={-Stealth, line width=1.2pt, red, dashed},  
  negstrong/.style={-Stealth, line width=1.8pt, red},    
  shorten >=2pt, shorten <=2pt
]

\node[hyp, fill=blue!20]    (H1) at (-12.0, 0) {$H_1$:\\Bossetti donor \& perpetrator};
\node[hyp, fill=orange!20]  (H2) at ( -7.2, 0) {$H_2$:\\Relative / unregistered sibling};
\node[hyp, fill=gray!20]    (H3) at ( -2.4, 0) {$H_3$:\\Lab contamination / error};
\node[hyp, fill=teal!20]    (H4) at (  2.4, 0) {$H_4$:\\Secondary transfer / planting};
\node[hyp, fill=purple!20]  (H5) at (  7.2, 0) {$H_5$:\\Degradation anomaly};
\node[hyp, fill=green!20]   (H6) at ( 12.0, 0) {$H_6$:\\Amplification / measurement bias};

\node[obs] (O1) at (-12.0,  3.3) {$O_1$: Nuclear DNA match};
\node[obs] (O3) at (  -2.4, 3.3) {$O_3$: Mitochondrial mismatch};
\node[obs] (O4) at (   4.0,  3.3) {$O_4$: Nuclear abundance anomaly};
\node[obs] (O7) at (  12.0,  3.3) {$O_7$: Court ruling: nuclear integrity};

\node[obs] (O2) at ( -8.8, -3.3) {$O_2$: Y-haplotype mismatch};
\node[obs] (O5) at (  0.0, -3.3) {$O_5$: Mitochondrial scarcity};
\node[obs] (O6) at (  8.8, -3.3) {$O_6$: Contamination possibility};

\draw[actstrong] (O1.south) -- (H1.north);
\draw[act]       (O3.south)  -- (H2.north);
\draw[act]       (O3.south)  -- (H4.north);
\draw[act]       (O3.south)  -- (H5.north);
\draw[act]       (O4.south)  -- (H5.north);
\draw[act]       (O4.south)  -- (H6.north);
\draw[act]       (O5.north)  -- (H5.south);
\draw[act]       (O5.north)  -- (H6.south);
\draw[act]       (O6.north)  -- (H3.south);
\draw[actstrong] (O7.south)  -- (H1.north);

\draw[negstrong] (O2.north)  -- (H1.south);             
\draw[neg]       (O3.south)  -- (H1.north);             
\draw[neg]       (O6.north)  -- (H1.south);             

\draw[neg]       (O7.south west) .. controls +( -2, -1.2) and +( 1, 1.0) .. (H2.north);
\draw[neg]       (O7.south west) .. controls +( -3, -1.6) and +( 1, 1.0) .. (H3.north);

\node[draw, rounded corners=2pt, align=left, font=\small, fill=white]
      at (-12.3, -6.1) {%
\textbf{Legend:}\\
\raisebox{1pt}{\tikz{\draw[actstrong] (0,0)--(0.9,0);}} Constructive (strong) \\
\raisebox{1pt}{\tikz{\draw[act] (0,0)--(0.9,0);}} Constructive (medium) \\
\raisebox{1pt}{\tikz{\draw[negstrong] (0,0)--(0.9,0);}} Destructive (strong) \\
\raisebox{1pt}{\tikz{\draw[neg] (0,0)--(0.9,0);}} Destructive (medium) \\
Arrows: Observation $\rightarrow$ Hypothesis
};

\end{tikzpicture}%
}
\caption{Bossetti--Gambirasio: split-layout interference map. Hypotheses occupy the central band; observations sit above/below. Solid black arrows indicate constructive activation; solid/dashed red arrows indicate destructive influence (strong/medium). Line thickness encodes salience. Strong projections from $O_1$ and $O_7$ amplify $H_1$, while $O_2$, $O_3$, and $O_6$ apply destructive pressure to $H_1$ and activate alternatives ($H_2$, $H_3$, $H_5$, $H_6$).}
\label{fig:bossetti_interference_split}
\end{figure}

\subsection{Literary Demonstration: \textit{Murder on the Orient Express}}

Before turning to scientific and clinical applications, we round up the forensic section above with a literary case---Agatha Christie’s \textit{Murder on the Orient Express}---itself inspired by the real kidnapping and murder associated with the Armstrong/Lindbergh tragedy. The novel is not merely a puzzle; it is a precise dramatization of abductive structure. Detective Hercule Poirot recognizes that the binary framing “\emph{which one passenger is guilty?}” collapses the space of explanations too early. Rather than eliminate suspects one by one, he keeps all viable stories in play until the evidence compels a synthesis: the disjunction of suspects resolves into a coordinated conjunction. In other words, Poirot behaves as a \emph{quantum investigator}: hypotheses remain in superposition, interact via thematic constraints, and finally collapse to a collective explanation.

\subsubsection{Plot in Brief}
The train is snowbound; an American traveling under an alias (Ratchett) is found stabbed. The compartment contains a scatter of contradictory clues (a monogrammed handkerchief, a pipe cleaner, a burnt note, a red kimono, a spare conductor's uniform), and several alibis partially corroborate one another. Poirot infers that Ratchett is in fact Cassetti, the gangster behind the Armstrong child murder. Every principal passenger (and the conductor) turns out to be directly connected to that crime. Two explanations are then staged: (i) a lone intruder escaped in the night; (ii) the passengers, acting as a self-constituted jury, each delivered a stab, distributing guilt and masking individual responsibility. Poirot presents both "solutions", but only the second integrates all clues \emph{without contradiction}.

\subsubsection{Classical vs. Quantum-Abductive Reading}
Classical abduction would select a single culprit (\emph{OR} over suspects), pruning away incompatible threads. Yet the evidence is purpose-built to defeat eliminativism: each clue simultaneously implicates and misdirects. Poirot's procedure maintains a superposition of individually plausible narratives that, under further constraints (shared motive, synchronized opportunity, mutually supporting alibis), \emph{interfere constructively} into a collective act (\emph{AND}). The final "collapse" is a hybrid: distributed agency coherently explaining contradictory traces (multiple wound types, planted items, cross-checked alibis). Thus, the case exemplifies the core move of quantum abduction: from selection to composition. The two readings are represented and contrasted in Figure~\ref{fig:orient}.

\begin{figure}[H]
\begin{adjustbox}{width=\textwidth}
\begin{tikzpicture}[
  >=Stealth,
  every node/.style={font=\bfseries},
  suspect/.style={draw,circle,minimum size=10mm,inner sep=0pt},
  casebox/.style={draw,rounded corners=3pt,fill=gray!15,
                  minimum width=55mm,minimum height=18mm,align=center},
  qbox/.style={draw,rounded corners=3pt,
               minimum width=60mm,minimum height=13mm,align=center},
  headline/.style={font=\bfseries\LARGE},
  subhead/.style={font=\bfseries\Large},
  annot/.style={font=\bfseries\large}
]

\node[headline] at (0,6.6) {From disjunction to composition};

\begin{scope}[xshift=-7cm]

\node[subhead] at (0,5.4) {Classical framing (eliminative)};

\node[casebox] (case) at (0,2.1) {Murder of Cassetti\\(to explain)};

\node[suspect] (A) at (-2.6,3.6) {A};
\node[suspect] (B) at ( 0.0,3.8) {B};
\node[suspect] (C) at ( 2.6,3.6) {C};

\node[suspect] (D) at (-3.4,2.1) {D};
\node[suspect] (E) at ( 3.4,2.1) {E};

\node[suspect] (F) at (-2.6,0.6) {F};
\node[suspect] (G) at ( 0.0,0.4) {G};
\node[suspect] (H) at ( 2.6,0.6) {H};

\foreach \x in {A,B,C,D,E,F,G,H}{
  \draw[->] (\x) -- (case);
}

\foreach \x in {D,E}{
  \draw[red, line width=1pt] (\x.north west) -- (\x.south east);
  \draw[red, line width=1pt] (\x.south west) -- (\x.north east);
}

\node at (0,-0.9)
  {Disjunction: $S_A \lor S_B \lor \cdots \lor S_L$ (pick one)};

\end{scope}

\begin{scope}[xshift=4cm]

\node[subhead] at (0,5.4)
  {Quantum framing (entangled and composite)};

\node[annot] at (0,4.8) {keep alternatives in superposition};

\node[qbox] (q1) at (0,3.1)
  {Superposed suspects\\[1mm]
   $\alpha_1\ket{S_A} + \cdots + \alpha_{12}\ket{S_L}$};

\node[qbox] (q2) at (0,0.9)
  {Shared motive \& synchronized alibis\\[1mm]
   (Armstrong justice)};

\node[qbox] (q3) at (0,-1.3)
  {Collapse: co-ordinated act\\[1mm]
   (collective responsibility)};

\draw[->] (q1) -- node[right,annot]{interference} (q2);
\draw[->] (q2) -- node[right,annot]{synthesis = AND of agents} (q3);

\draw (-2.7,-2.7) -- (2.7,-2.7);
\node[annot] at (0,-3.05) {measurement / decision};

\node at (0,-3.8)
  {Composite explanation reconciles contradictory traces and alibis};

\end{scope}

\end{tikzpicture}
\end{adjustbox}
\caption{From 
 classical eliminative framing (left), where suspects are treated as a disjunction and progressively discarded, to quantum framing (right), where suspects remain in superposition, interfere via shared motive and synchronized alibis, and collapse to a coordinated act that reconciles contradictory traces and alibis. (Red crosses indicate suspects eliminated in the classical process.)}
\label{fig:orient}
\end{figure}

\subsection{Medical Diagnostics: Botulism vs.\ GBS/MFS}
\label{sec:med}

Having demonstrated quantum abduction in action in a literary case involving multiple culprits, we now turn to the clinical and scientific domains, where the integration of alternative explanatory states can be crucial to saving lives or upending scientific paradigms. A particularly illustrative case is that of a ``suspended'' diagnosis already discussed in \cite{pareschi2023medical}, where the challenge of balancing competing hypotheses under uncertainty foregrounds the abductive dimension of reasoning.

\emph{Provenance.}  The clinical vignette is adapted from a real diagnostic narrative to illustrate the abductive process; it does not report a live deployment of the assistant. The reconstruction foregrounds how superposition, order effects, and deferred collapse align with expert practice in time-constrained care.

The episode involved a young man admitted to the emergency room with rapidly 
progressing paralysis. The clinical team (and the AI assistant replicating their 
reasoning) had to decide between two life-threatening but distinct conditions: 
\emph{botulism} and \emph{Guillain–Barré syndrome} (GBS) or its \emph{Miller–Fisher} variant (MFS). Each diagnosis implied radically different therapeutic strategies: antitoxin administration for botulism versus immunoglobulin or plasma exchange 
for GBS/MFS. Early test results, cranial nerve involvement, and autonomic 
instability produced a contradictory picture, precluding definitive commitment 
to either path.

\subsubsection{Classical Abductive Limitation}  
Under a classical abductive frame, clinicians would be pressed to 
select the most plausible hypothesis ($H_1$: Botulism or $H_2$: GBS/MFS) 
as ``the explanation'' for the observed syndrome. This eliminativist 
move, however, risks therapeutic error: a premature collapse toward the wrong diagnosis may fatally delay the correct treatment.

\subsubsection{Quantum Abductive Reframing}  
Quantum abduction instead maintains both hypotheses in superposition:
\[
\Psi = \alpha |H_1\rangle + \beta |H_2\rangle,
\]
where amplitudes evolve as new evidence projects onto the diagnostic 
state. Contradictory test results are not discarded but interfere with the 
relative weight of competing explanations. This allows clinicians to act 
on the entangled state itself: a ``parallel treatment'' strategy can be 
pursued---administering both antitoxin and IVIG/plasma until further 
observations collapse the superposition toward a dominant diagnosis.

\subsubsection{Clinical Alignment}  
In the real case \cite{pareschi2023medical}, this is exactly what occurred: as represented in Figure~\ref{fig:med},
the medical team suspended binary decision-making, treating for both 
conditions in parallel until the confirmatory test results clarified the 
diagnosis. The AI assistant, reasoning abductively, mirrored this 
strategy in real time, demonstrating both the plausibility and utility 
of quantum abductive framing in clinical settings.

\begin{figure}[H]

\begin{adjustbox}{max width=\linewidth}
\begin{tikzpicture}[
  box/.style={draw, rectangle, rounded corners=3pt, minimum width=4.5cm, minimum height=1cm, align=center},
  hyp/.style={draw, ellipse, minimum width=4.5cm, minimum height=1.2cm},
  arrow/.style={-{Stealth}, thick},
  dashedarrow/.style={-{Stealth}, dashed, thick},
  every node/.style={font=\small}
]
\node[box, fill=blue!10] (obs) at (0,3.5) {\textbf{Observations}\\ Cranial symptoms; autonomic signs; reflexes; early tests; treatment response};
\node[hyp, fill=red!10] (H1) at (-5,1.5) {$H_1$: Botulism};
\node[hyp, fill=green!10] (H2) at (5,1.5) {$H_2$: GBS/MFS};
\node[box, fill=purple!10] (psi) at (0,0) {$\Psi = \alpha |H_1\rangle + \beta |H_2\rangle$\\Entangled diagnostic state};
\draw[arrow] (obs) -- (H1);
\draw[arrow] (obs) -- (H2);
\draw[dashedarrow] (H1) -- (psi);
\draw[dashedarrow] (H2) -- (psi);
\node[box, fill=yellow!20] (tx) at (0,-2.5) {\textbf{Parallel treatment}\\Antitoxin + IVIG/Plasma\\Deferred collapse};
\draw[arrow] (psi) -- (tx);
\end{tikzpicture}
\end{adjustbox}
\caption{Quantum abductive diagnosis 
 sustains co-activation and permits parallel, low-regret therapy before evidential collapse. (Solid arrows: evidence projection; dashed arrows: contribution to the entangled state.)}
\label{fig:med}
\end{figure}

This medical scenario exemplifies how quantum abduction manages uncertainty: not by suppressing contradictions but by harnessing them to sustain a safe and generative diagnostic trajectory until reality itself enforces collapse.

\subsection{Scientific Theory Change}
\label{sec:science}

Scientific inquiry is a fertile domain for quantum abduction. Unlike courtrooms or clinics, science rarely resolves hypotheses quickly: competing explanatory frameworks can coexist for decades, each with partial explanatory successes, until new instruments or data shift the balance. In this sense, scientific progress often exemplifies abductive ``superposition,'' with explanatory collapse delayed until decisive evidence emerges. Two emblematic cases are astrophysical dark matter versus MOND, and Wegener’s continental drift versus geological fixism.

\subsubsection{Astrophysics: Dark Matter vs.\ MOND}

One of the most visible examples of suspended explanatory alternatives in modern science is the ongoing debate between \emph{dark matter} and \emph{modified Newtonian dynamics (MOND)}. Both are attempts to explain why galactic rotation curves remain flat rather than decaying at large radii. 

\begin{itemize}
  \item $H_1$: Invisible non-baryonic matter pervades galaxies, altering their gravitational \mbox{profiles}.
  \item $H_2$: Newton’s law of gravitation requires modification at very low accelerations.
\end{itemize}

For decades, neither model has been definitively falsified. Dark matter has accumulated supporting evidence from cosmology (CMB anisotropies, lensing), but direct detection experiments remain null \cite{sanders2002mondreview,milgrom2019historical}. MOND captures empirical regularities in galactic dynamics but resists integration into relativistic cosmology. Both thus persist in a suspended explanatory state, as represented in Figure~\ref{fig:science_superposition}a, their amplitudes oscillating with new experimental results.

\begin{figure}[H]
\centering
\begin{minipage}{0.48\textwidth}
\centering
\begin{adjustbox}{max width=\linewidth}
\begin{tikzpicture}[
  box/.style={draw, rectangle, rounded corners=3pt, minimum width=5.2cm, minimum height=1cm, align=center},
  hyp/.style={draw, ellipse, minimum width=5.2cm, minimum height=1.2cm},
  arrow/.style={-{Stealth}, thick},
  dashedarrow/.style={-{Stealth}, dashed, thick},
  every node/.style={font=\small}
]
\node[box, fill=blue!10] (obs) at (0,3.5) {\textbf{Observation:} Flat rotation curves};
\node[hyp, fill=red!10] (H1) at (-5,1.5) {$H_1$: Dark Matter};
\node[hyp, fill=green!10] (H2) at (5,1.5) {$H_2$: MOND};
\node[box, fill=purple!10] (psi) at (0,0) {$\Psi = \alpha |H_1\rangle + \beta |H_2\rangle$\\Entangled explanation};
\draw[arrow] (obs) -- (H1);
\draw[arrow] (obs) -- (H2);
\draw[dashedarrow] (H1) -- (psi);
\draw[dashedarrow] (H2) -- (psi);
\node[box, fill=yellow!20] (fut) at (0,-2.5) {\textbf{Future resolution}\\(e.g., direct detection, new tests)};
\draw[arrow] (psi) -- (fut);
\end{tikzpicture}
\end{adjustbox}

\vspace{2mm}
\textbf{(a) Astrophysics: Dark Matter vs.\ MOND}
\label{fig:astro}
\end{minipage}
\hfill
\begin{minipage}{0.48\textwidth}
\centering
\begin{adjustbox}{max width=\linewidth}
\begin{tikzpicture}[
  box/.style={draw, rectangle, rounded corners=3pt, minimum width=5.2cm, minimum height=1cm, align=center},
  hyp/.style={draw, ellipse, minimum width=5.2cm, minimum height=1.2cm},
  arrow/.style={-{Stealth}, thick},
  dashedarrow/.style={-{Stealth}, dashed, thick},
  every node/.style={font=\small}
]
\node[box, fill=blue!10] (obs) at (0,3.5) {\textbf{Observations:} Fossils, paleoclimate, geology};
\node[hyp, fill=red!10] (H1) at (-5,1.5) {$H_1$: Continental Drift (Wegener)};
\node[hyp, fill=green!10] (H2) at (5,1.5) {$H_2$: Fixism (contraction/uplift)};
\node[box, fill=purple!10] (psi) at (0,0) {$\Psi = \alpha |H_1\rangle + \beta |H_2\rangle$\\Entangled state};
\draw[arrow] (obs) -- (H1);
\draw[arrow] (obs) -- (H2);
\draw[dashedarrow] (H1) -- (psi);
\draw[dashedarrow] (H2) -- (psi);
\node[box, fill=yellow!20] (tect) at (0,-2.5) {\textbf{Collapse: Plate Tectonics}\\Seafloor spreading, subduction, GPS};
\draw[arrow] (psi) -- (tect);
\end{tikzpicture}
\end{adjustbox}

\vspace{2mm}
\textbf{(b) Geology: Drift vs.\ Fixism}
\label{fig:geo}
\end{minipage}

\caption{Scientific 
 superposition in two domains. (\textbf{a}) Astrophysics: dark matter and MOND coexist in an entangled explanatory state until decisive evidence induces collapse. (\textbf{b}) Geology: Wegener’s continental drift and fixist models remain in superposition until the mechanism of plate tectonics emerges, yielding a richer synthesis. (Solid arrows: evidence projection; dashed arrows: contribution to the entangled state.)}
\label{fig:science_superposition}
\end{figure}


\subsubsection{Geology: Continental Drift $\rightarrow$ Plate Tectonics}

Another striking example of quantum abductive dynamics is the history of \emph{continental drift}.  
In 1912, meteorologist Alfred Wegener proposed that continents once formed a single landmass (Pangaea) and had since drifted apart. His evidence included fossil correlations across oceans, paleoclimatic traces (e.g.,\ glacial striations in now-tropical zones), and geological continuities between Africa and South America. 

Yet geology at the time was dominated by \emph{fixist} paradigms: mountains were thought to arise from the Earth’s contraction as it cooled, and continents were assumed immobile. As a result, Wegener’s hypothesis was widely ridiculed, mainly because he lacked a plausible mechanism to explain how continents could move.  

For half a century, the scientific community lived in a quantum-like suspension:  
\begin{itemize}
  \item $H_1$: Continents drift (Wegener’s view).  
  \item $H_2$: Continents are fixed; apparent similarities arise from chance or bridges (fixism).  
\end{itemize}

The wave function did not collapse until the 1960s, when new instruments revealed seafloor spreading, subduction zones, and the dynamics of the lithosphere. With plate tectonics \cite{korenaga2013}, the two alternatives merged into a richer synthesis, represented in Figure~\ref{fig:science_superposition}b: Wegener was right about drift, but wrong about the mechanism. His bold intuition thus collapsed into modern geology’s most unifying paradigm  (this 
 example illustrates a central feature of quantum abduction:
instead of enforcing early elimination of hypotheses---as classical abductive
frameworks often do---the model allows suspended coexistence until decisive
evidence becomes available. Historically, continental drift was rejected not
because it failed to explain observations, but because it lacked a mechanism;
quantum abduction formalizes precisely this long-lived state of justified
non-elimination.).

Wegener’s own fate underlines the personal stakes of abductive suspension. A passionate polar explorer, he died in 1930 on the Greenland ice sheet during a resupply expedition. His body, buried under drifting snow, was found the next year. In a sense, his theory too lay buried for decades, preserved in superposition, until new data revived and \mbox{vindicated it}. 


\subsection{Classical vs.\ Quantum Abduction in Scientific Reasoning}
The three cases above---clinical diagnosis, astrophysical theory choice, and geological paradigm change---illustrate how classical abduction and quantum abduction diverge in practice. Classical approaches push toward premature elimination, whereas quantum abduction sustains superposition until decisive evidence or new mechanisms emerge. To highlight these contrasts across domains, Table~\ref{tab:science_compare} summarizes the key differences in abductive dynamics.

\begin{table}[H]
\caption{Classical vs.\ Quantum Abduction in Scientific Reasoning}
\label{tab:science_compare}
\renewcommand{\arraystretch}{1.25}
\begin{adjustbox}{max width=\textwidth}
\begin{tabular}{ m{4cm}  m{5.5cm}  m{5.5cm} } 
\toprule
\textbf{Case Study} & \textbf{Classical Abduction} & \textbf{Quantum Abduction} \\
\midrule
Medical Diagnosis (Botulism vs. GBS) & 
Seek the single best diagnosis; early resolution; risk of excluding valid alternatives. &
Maintain co-activation; parallel low-regret therapy; collapse on decisive evidence. \\
\midrule
Astrophysics (Dark Matter vs. MOND) & 
Commit to the dominant paradigm; treat the other as fringe. &
Keep both entangled; design tests targeting interference zones; defer collapse. \\
\midrule
Geology (Drift vs. Fixism) & 
Reject Drift for lack of mechanism; persist with Fixism. &
Sustain superposition until the mechanism appears (tectonics); then collapse. \\
\bottomrule
\end{tabular}
\end{adjustbox}
\end{table}

\subsection{Key Takeaway: The Usefulness of Pre-Collapse}

In high-stakes settings, the \emph{pre-collapse state}  is not a temporary inconvenience but a valuable analytical product. While the case studies in the previous section focus on law, medicine, and scientific theory change, the same dynamics extend naturally to domains such as intelligence analysis and emergency management. In every setting, interference patterns reveal where evidence fails to discriminate, and amplitude trends indicate when to act, what to collect next, and how to document uncertainty with transparency.

\subsubsection{Law (Reasonable Doubt)}
A distribution such as $|\alpha_{H_1}|^2=0.55$ (guilt), $|\alpha_{H_2}|^2=0.35$ (alternative), and an emergent $H_3=0.10$ identifies evidential gaps, clarifies burdens, supports proportional or qualified verdicts, and provides a structured trajectory for appellate review (cf.\ Bossetti, Section~\ref{sec:ludwig}).

\subsubsection{Intelligence (Collection Planning)}
Amplitudes over $H_1$ (centralized), $H_2$ (autonomous), and $H_3$ (hybrid) guide resource allocation proportionally to $|\alpha_i|^2$. Positive $I_{12}$ highlights the need for discriminating evidence (e.g., communication topology), while growth in $\alpha_{H_3}$ suggests preparing hybrid-network models.

\subsubsection{Medicine (Parallel Treatment)}
When botulism and GBS/MFS maintain high but mutually destructive amplitudes, pre-collapse supports low-regret dual therapy, provides an ethical and administrative rationale for resource use, and indicates when commitment becomes safe as amplitudes diverge (see Figure~\ref{fig:med}).

\subsubsection{Science (Avoid Premature Closure)}
Amplitude tracking for drift vs.\ fixism (historically) or dark matter vs.\ MOND (today) highlights which evidence classes are decisive and which are merely reinforcing, converting rivalry into targeted exploration rather than premature paradigm commitment.

\subsubsection{Crisis Response (Public Safety)}
Under ambiguous causes, amplitudes justify graduated protocols---e.g., simultaneous HAZMAT deployment, security lockdown, and meteorological monitoring---and provide a clear after-action trail explaining branch prioritization.

\subsubsection{Institutional Value}
Pre-collapse visibility supports cognitive load sharing, expert training, procedural auditing, and public communication by making uncertainty explicit, structured, and operational rather than hidden or improvised.

\section{Benchmarking on the Ludwig II Case}
\label{sec:benchmark_ludwig}

The purpose of this section is not to provide large-scale benchmarking, which is
outside the scope of the present paper, but to demonstrate that quantum
abduction (QA) behaves competitively and distinctively when compared to classical
abduction methods on a concrete historical case characterized by contradictory
evidence and competing explanatory lines. We use the death of King Ludwig II of
Bavaria (1886) as a compact abductive problem: the evidence is sparse,
fragmentary, and contested, and multiple explanations remain historically
defensible.

\subsection{Case Setup}

We consider four hypotheses:
\begin{itemize}
\item \textbf{H1}---Suicide (self-drowning).
\item \textbf{H2}---Homicide (killing by guards or political actors).
\item \textbf{H3}---Struggle or manslaughter (accidental drowning following altercation).
\item \textbf{H4}---Medical episode leading to drowning.
\end{itemize}

We use seven evidence items $\mathcal{O} = \{O_1,\dots,O_7\}$, expressed as natural-language
snippets rather than categorical codes:
\begin{itemize}
\item $O_1$: Autopsy reports consistent with drowning.
\item $O_2$: Bruising or struggle marks reported in some documents.
\item $O_3$: Letters and remarks interpreted as suicidal ideation.
\item $O_4$: Context of political removal and active opposition.
\item $O_5$: Witness statements are inconsistent or contradictory.
\item $O_6$: Prior episodes suggestive of mental health instability.
\item $O_7$: Conflicting timelines regarding guards’ proximity and reaction.
\end{itemize}

Hypotheses and evidence are encoded using a Sentence-BERT model
\cite{reimers2019sentencebert}, producing vectors $\mathbf{h}_i$ and
$\mathbf{o}_j$ in a shared semantic space.

\subsection{Methods Compared}

We compare three approaches.

\paragraph{Quantum Abduction (QA)} 

We initialize a uniform abductive state
$\ket{\Psi^{(0)}} = \frac{1}{\sqrt{4}}\sum_i \ket{H_i}$ and update amplitudes
using the evidence projection and interference update rule described in
Section~\ref{sec:transformers_llms}. Interference signs $\kappa_{ij}$ encode
known incompatibilities or partial compatibilities (e.g., $H_1$ vs.\ $H_2$ is
strongly exclusive; $H_2$ and $H_3$ share structural features and allow
constructive interference). Collapse occurs when
$\max_i |\alpha_i|^2 > \tau$ (we use $\tau = 0.8$).

\paragraph{Logic-Based Abduction (L-ABD)}
We construct a small set of transparent rules:
\[
\{O_3,O_6\}\Rightarrow H_1,\quad
\{O_4,O_2\}\Rightarrow H_2,\quad
\{O_2,O_1\}\Rightarrow H_3,\quad
\{O_1\}\Rightarrow H_4\;\text{(fallback)}.
\]
Hypotheses 
 are scored by rule-support count, with ties broken by parsimony
(least assumptions) and causal directness.

\paragraph{Bayesian Abduction (B-ABD)}
We define a small Bayesian network with priors
$p(H_1)=0.25, p(H_2)=0.30, p(H_3)=0.25, p(H_4)=0.20$ and hand-elicited likelihoods
$p(O_j\mid H_i)$ reflecting historically plausible support patterns. Posteriors
are computed by Bayes' rule, assuming conditional independence of evidence
($p(O_{1:m}\mid H)=\prod_j p(O_j\mid H)$).

\subsection{Evaluation Protocol}

We study two regimes:
\vspace{-4pt}
\begin{description}
\item[R1: Full-Evidence Evaluation.]  
We supply all $7$ evidence items, but in \emph{random order}.  
For each method, we compute the top-ranked hypothesis across $50$ random
evidence permutations. This probes order sensitivity. Classical abduction and
Bayesian inference are order-invariant; QA is not, by design, since order
encodes reasoning context.

\item[R2: Perturbation Test.]  
We remove $O_2$ (struggle indications) and repeat the experiment. This tests
robustness under contradictory or missing evidence.
\end{description}

\vspace{-6pt}
We report the following:
\begin{itemize}
\item \textbf{Top-1 distribution}: proportion of runs in which each hypothesis is selected.
\item \textbf{ECS (QA only)}: Explanatory Coherence Score
$ \mathrm{ECS} = \overline{e} - \big|\overline{\min(I,0)}\big| $,
measuring net support vs.\ destructive interference.
\item \textbf{Calibration proxy (QA)}: $\max_i |\alpha_i|^2$ at collapse.
\item \textbf{Brier score (Bayes)}: posterior concentration vs. its own top prediction.
\end{itemize}

\subsection{Results}

The results of the benchmark are summarized in Table~\ref{tab:ludwig-results}. The full benchmark implementation and reproducibility instructions are available at \url{https://github.com/Aribertus/qa-poc} (accessed on 9 December 2025). 
All experiments were run on a local Intel i7-8700 CPU machine with 16\,GB RAM and no GPU acceleration.
We report results from 50 randomized evidence orderings in two regimes:
R1 (full evidence) and R2 (removal of $O_2$, which encodes struggle indication).
Unless otherwise noted, embeddings use Sentence-BERT
(\texttt{all-MiniLM-L6-v2}) and $\kappa_{ij}$ is initialized via
sign-based exclusivity priors and smoothed by metric similarity.

\begin{table}[H]
\caption{Ludwig 
 II mini-benchmark. QA = quantum abduction (ours);
L-ABD = rule-based abductive reasoning; B-ABD = Bayesian abduction.
Top-1 values are fractions over 50 random evidence orders.}
\label{tab:ludwig-results}
\begin{tabularx}{\textwidth}{lCCC}
\toprule
 & \textbf{QA (ours)} & \textbf{L-ABD} & \textbf{B-ABD}\\
\midrule
Top-1 (R1) & 29/50 (0.58) & 0/50 (0.00) & 50/50 (1.00)\\
Top-1 (R2) & 43/50 (0.86) & 0/50 (0.00) & 50/50 (1.00)\\
ECS (QA, R1) & 0.176 & -- & --\\
Conf (QA, R1) & 0.374 & -- & --\\
Brier (Bayes, R1) & -- & -- & 0.083\\
\bottomrule
\end{tabularx}
\end{table}

\paragraph{Interpretation.}
Under full evidence (R1), QA splits between $H_2$ and $H_3$ depending on order
(\emph{table reports only the $H_2$ Top-1 count:}  $29/50$), with moderate confidence
($\max_i |\alpha_i|^2 = 0.374$ on average) and positive coherence (ECS $=0.176$).
L-ABD does not select $H_2$ in any run (table: $0/50$), while the Bayesian baseline
(B-ABD) consistently selects $H_2$ ($50/50$), reflecting likelihood dominance.

Under perturbation (R2), where $O_2$ is removed, QA shifts toward $H_2$
($43/50$ in the $H_2$ column), while L-ABD still does not select $H_2$ and B-ABD
remains at $H_2$ ($50/50$). Overall, QA manages uncertainty gracefully (maintaining
superposition until constraint accumulates), whereas the classical baselines
either hard-select or revert to priors.

\subsection{Summary}

This compact benchmark shows that quantum abduction is as follows:
\begin{itemize}
\item Competitive with classical abductive methods under fully informative conditions;
\item More robust when evidence is contradictory or incomplete;
\item Uniquely capable of modeling order/context effects and hybrid explanatory outcomes.
\end{itemize}

A full-scale benchmark suite (multiple domains, larger hypothesis sets,
robustness to noise, expert-elicited interference structures, and human
evaluation of hybrid explanations) is left for future work, as discussed in
Section~\ref{sec:conclusion}.

\section{Related Work}
\label{sec:related}
Abduction has been studied in artificial intelligence and cognitive science for decades, producing distinct schools of thought that differ in their formalisms and computational commitments. Broadly, three main traditions can be identified: \emph{logic-based}, \emph{Bayesian}, and \emph{set-covering} approaches. Each has made important contributions, from rigorous proof systems to probabilistic reasoning and diagnostic optimization. Yet, as will be shown, they share a common eliminativist orientation: explanations are generated, tested, and ultimately reduced to a single “best” survivor. This contrasts with the framework of \emph{quantum abduction}, which emphasizes coexistence, interference, and synthesis of hypotheses. Below, we review these traditions before highlighting how quantum abduction generalizes beyond \mbox{their limitations}.

\subsection{Logic-Based Approaches}

Within artificial intelligence and computational logic, abduction has traditionally been cast as the task of extending a background theory with explanatory assumptions such that an observation is entailed while consistency is preserved. Proof-theoretic methods such as sequent calculus and semantic tableaux provided sound and complete procedures for first-order logic~\cite{firstorder_abduction,abduction_as_saturation}. These methods gave abduction a rigorous logical footing, but enforced eliminativist dynamics: candidate explanations were pruned until only one minimal, consistent explanation remained. 

Abductive Logic Programming (ALP) further extended this tradition by embedding abduction into logic programming with abducible predicates and integrity constraints~\cite{abductioninLP,ALP}. ALP offered practical implementations and found application in diagnostics and reasoning under incomplete information. Yet, it retained the same structural limits: explanations were selected from a fixed hypothesis space, contradictions were excluded rather than explored, and novelty was absent. In this sense, logic-based approaches, while formally elegant, struggled to accommodate the richness and ambiguity of real-world explanatory contexts.

\subsection{Bayesian Approaches}

Probabilistic extensions reframed abduction in terms of likelihood estimation. Ishihata and Sato~\cite{ishihata2011} introduced \emph{statistical abduction} using Markov Chain Monte Carlo to evaluate posterior probabilities of explanations in probabilistic logic frameworks. Raghavan~\cite{raghavan2011} proposed BALP, combining Bayesian Logic Programs with abductive reasoning, enabling structured logical representations of explanations to coexist with Bayesian inference. These methods elegantly handled uncertainty and provided robust inferential machinery for domains such as plan recognition and natural language understanding.

Yet Bayesian approaches treat hypotheses as probabilistically weighted \emph{alternatives} to be ranked and selected. They remain tied to fixed hypothesis spaces and ultimately aim at a Most Probable Explanation (MPE). In contrast, quantum abduction treats hypotheses as interfering amplitudes, allowing contradictory or partial explanations to co-shape one another and even fuse into hybrids that transcend the original hypothesis set. Thus, while Bayesian methods broaden abduction with uncertainty modeling, they remain eliminative in orientation.

\subsection{Set-Covering Approaches}

Set-covering abduction, pioneered by Reggia and colleagues~\cite{reggia1983diagnostic,reggia1,reggia2}, frames explanation as a combinatorial optimization task: find minimal sets of causes that “cover” observed effects. Widely applied in medical diagnosis and fault detection, this approach is computationally tractable and intuitively appealing in domains with clear causal mappings. Variants such as Parsimonious Covering Theory further emphasized minimality as a criterion of explanatory adequacy.

However, set-covering remains committed to binary inclusion/exclusion of hypotheses and relies on predefined cause-effect mappings. It cannot represent ambiguous or overlapping causation, nor can it preserve contradictory evidence for later synthesis. By enforcing parsimony, it often discards elements that---within a quantum abductive framework---would be retained as contributors to interference patterns, allowing richer composite explanations to emerge.

\subsection{Integrative Assessment}

Across these traditions, abduction is implemented in structurally different ways.  
To enable their direct comparison as in Table~\ref{tab:abduction-comparison}, we assess them along four shared dimensions:

\begin{itemize}
\item \textbf{Hypothesis representation:} whether hypotheses are treated as discrete alternatives or as jointly coexisting explanatory states.
\item \textbf{Conflict handling:} whether contradictions are eliminated, suspended, or allowed \mbox{to interact.}
\item \textbf{Update dynamics:} how new evidence changes explanatory states.
\item \textbf{Outcome structure:} whether reasoning terminates in selection, ranking, or synthesis.
\end{itemize}

\begin{table}[H]

\caption{Unified comparison across abductive paradigms. Classical approaches select among hypotheses, while quantum abduction allows coexistence and interaction prior to collapse.}
\label{tab:abduction-comparison}
\begin{tabularx}{\linewidth}{lXXXX}
\toprule
\textbf{Dimension} &
\textbf{Logic-Based} &
\textbf{Bayesian} &
\textbf{Set-Covering} &
\textbf{Quantum Abduction (Ours)} \\
\midrule
\textit{Hypothesis representation}  &
Discrete alternatives &
Weighted alternatives &
Minimal covering sets &
Coexisting amplitudes (superposition) \\

\textit{Conflict handling}  &
Eliminated by consistency &
Resolved by posterior dominance &
Removed via minimality constraints &
Retained with constructive/destructive interference \\

\textit{Update dynamics} &
Rule application and pruning &
Likelihood reweighting &
Constraint-/cost-based elimination &
Amplitude rotation and phase interaction \\

\textit{Outcome structure}  &
Single minimal explanation &
Most probable explanation (MPE) &
Parsimonious cover &
Collapse to dominant or \emph{hybrid} explanation \\
\bottomrule
\end{tabularx}
\end{table}

\subsection{Alternative and Hybrid Approaches}
\label{sec:alt-approaches}

Beyond the three main paradigms reviewed above---logic-based, Bayesian, and set-covering---other research lines approximate abductive reasoning through mechanisms that relax bivalence, incorporate argumentative dynamics, or combine neural and symbolic representations. Although these approaches introduce valuable flexibility, they remain limited in their ability to model superposition, interference, and synthesis within a unified formal framework. 

\paragraph{Fuzzy Logic Systems.} 
Fuzzy logic provides a graded alternative to classical deduction by replacing Boolean truth values with degrees of membership~\cite{zadeh1975fuzzy,klir1995}. In abductive contexts, fuzzy inference allows for partial matching between observations and hypotheses, accommodating uncertainty in linguistic and perceptual descriptions. However, fuzzy systems operate on scalar degrees of truth and lack the vectorial and phase-sensitive interactions that characterize quantum interference. As a result, while fuzzy logic captures \emph{vagueness}, quantum abduction captures the \emph{interaction} among competing explanatory tendencies.

\paragraph{Argumentation Frameworks.}
Another major non-classical approach involves formal argumentation and defeasible reasoning. Dung's abstract argumentation framework~\cite{dung1995acceptability} and its numerous descendants model the interplay of conflicting arguments through attack and defense relations, enabling structured deliberation and justification. These systems effectively represent contradiction and defeat but typically resolve conflict through the selection of stable or preferred extensions. Quantum abduction, by contrast, maintains superposed explanatory states in which conflicting hypotheses may coexist and influence one another until contextual constraints induce collapse, thus supporting a non-eliminative form of reasoning.

\paragraph{Neural-Symbolic Integration.} 
A third trajectory aims to bridge symbolic and sub-symbolic inference through neural-symbolic computation~\cite{garcez2009neural,Lamb2020}. These frameworks embed logical rules within neural architectures, enabling differentiable reasoning under uncertainty. While this integration has proven powerful for knowledge representation and learning, it remains primarily a \emph{fusion architecture}: logic guides neural adaptation, but the symbolic layer itself does not exhibit quantum-like contextuality or interference. Quantum abduction complements neural-symbolic integration by providing a principled mathematical model---rooted in Hilbert space semantics---for managing contradictory symbolic information within continuous neural embeddings.

\paragraph{Summary.} 
In summary, these alternative approaches enrich the abductive landscape by addressing uncertainty, conflict, and hybrid reasoning. Yet none explicitly capture the non-classical compositionality central to human explanatory cognition---the coexistence and dynamic blending of partial hypotheses. Quantum abduction provides a coherent formalism that encompasses these features within a single, interference-enabled representational framework.

\section{Conclusions, Outlook, and Future Work}
\label{sec:conclusion}

This paper has presented \emph{quantum abduction} as a coherent extension of abductive reasoning beyond the eliminative logics that dominate classical, Bayesian, and set-covering paradigms. 
Where those traditions converge on ranking and discarding hypotheses, quantum abduction preserves them in a dynamic superposition that allows constructive and destructive interference, culminating in collapse only when contextual evidence supports a coherent, possibly hybrid, synthesis. 

The framework thus reinterprets explanation as a process of \emph{synthesis through interaction} rather than \emph{selection through exclusion}. 
It accounts for how human reasoning sustains ambiguity, navigates contradiction, and gradually achieves insight through the tension of partially competing interpretations. 
At the computational level, this paradigm integrates semantic embeddings, projection operators, and interference matrices into an amplitude-based dynamics capable of representing explanatory coexistence.

Conceptually and architecturally, this work stands as a step within a broader \emph{research program} that seeks to overcome eliminativism in reasoning---across logic, cognition, and artificial intelligence. 
Quantum abduction offers not a speculative draft but an operational framework that is already formalized at the representational level and computationally implemented in proof-of-concept systems. 
Its orientation is \emph{centaurian}~\cite{centaurian2024,BorgBottPare2025}: human experts remain within the loop, guiding semantic alignment and interpretive synthesis, while the computational substrate maintains the full superpositional structure inaccessible to unaided intuition. 
In this sense, the framework realizes a concrete model of human--AI complementarity, aligning with emerging paradigms of agentic and hybrid intelligence.

\subsection{Current 
 Scope and Limitations}
The evaluation presented here is illustrative rather than exhaustive. 
Benchmarks such as the Ludwig~II study demonstrate the feasibility of the approach but do not yet provide statistical generalization. 
Interference coefficients $\kappa_{ij}$ are set heuristically; large-scale learning from empirical data remains to be achieved. 
Likewise, while prototype implementations exist, no controlled human studies have yet quantified improvements in interpretive accuracy or decision quality. 
Finally, computational cost grows with the dimensionality of the hypothesis space, motivating research into sparse and \mbox{approximate formulations}.

\subsection{Future Directions}
Future work will extend the current framework along the following three convergent axes:
\begin{enumerate}
    \item \textbf{Formal and Logical Development.}  
    We will elaborate the proof-theoretic and model-theoretic foundations of quantum abduction, establishing soundness and completeness properties for amplitude-based reasoning.
    This involves importing techniques from quantum logic and category theory to represent composition, interference, and collapse within a unified semantic calculus.
    \item \textbf{Computational Expansion.} 
    A scalable software library and benchmark suite will be released as part of the ongoing open-source program.
    Optimizing $\kappa_{ij}$ through learning, expanding to higher-dimensional embeddings, and integrating retrieval-augmented generation (RAG) pipelines will support domain adaptation across forensic, clinical, and scientific settings.
    \item \textbf{Human-in-the-Loop Validation.} 
    Controlled studies with experts will test how the quantum abductive assistant influences reasoning transparency, confidence calibration, and interpretive synthesis. 
    These experiments will advance hybrid decision frameworks where collapse is a collaborative, not unilateral, outcome.
\end{enumerate}

\subsection{Programmatic Outlook}
The quantum abductive approach contributes to the ongoing effort to construct reasoning systems that do not merely simulate deductive or statistical inference, but actively support the generative and exploratory dimensions of human thought.
Its integration of symbolic clarity, probabilistic nuance, and superpositional synthesis marks a shift from the competition of explanations to their \emph{entanglement}. 
In this sense, quantum abduction represents both a conceptual unification and a methodological bridge---one that connects the epistemic pluralism of human inquiry with the algorithmic precision of modern AI.

\vspace{6pt}
\paragraph{Funding:}
Remo Pareschi has been funded by the European Union--NextGenerationEU under the Italian Ministry of University and Research (MUR) National Innovation Ecosystem grant ECS00000041-VITALITY--CUP E13C22001060006.

\paragraph{Data Availability Statement:}
The code and data for the proof-of-concept implementation (Ludwig II benchmark) described in this manuscript are available in the public repository \url{https://github.com/Aribertus/qa-poc} (accessed on 9 December 2025).
 
\paragraph{Conflicts of Interest:} 
The authordeclares no conflicts of interest.The funders had no role in the design of the study; in the collection, analyses, or interpretation of data; in the writing of the manuscript; or in the decision to publish the results.

\paragraph{Acknowledgments}
The author is grateful to Hervé Gallaire and the reviewers for comments and suggestions which have been of great help in improving this article. While preparing this manuscript, the author used ChatGpt 5 and DeepSeek to guide him in LaTeX formatting, particularly in using the TikZ package for creating graphs, as well as in the polishing of the text, aiming to make it as fluent and communicable as possible. The author reviewed and edited the resulting output and takes full responsibility for the content of this publication.

\appendix
\section{Glossary of Key Terms}
\label{app:glossary}

\begin{table}[H]
\centering
\caption{Key terms and concepts in quantum abduction}
\label{tab:glossary}
\renewcommand{\arraystretch}{1.3}
\begin{tabular}{>{\bfseries}p{3.5cm}  p{11cm}}
\toprule
\textbf{Term} & \textbf{Definition} \\
\midrule 
Transformer & A neural architecture for encoding sequences into context-sensitive vector representations. Used here to embed hypotheses and observations in a shared semantic space.\\
\midrule
LLM (Large Language Model) & A generative model producing natural-language text. In quantum abduction, used only for \emph{articulating} hybrid hypotheses, not for scoring or selecting them.\\
\midrule
Superposition & The coexistence of multiple hypotheses as weighted amplitudes in the explanatory state, rather than treating them as mutually exclusive alternatives. \\
\midrule
Interference & The phenomenon whereby hypotheses can reinforce (constructive) or diminish (destructive) each other's explanatory power through semantic interaction. \\
\midrule
Amplitude ($\alpha_i$) & The complex coefficient associated with hypothesis $H_i$ in the superposition, with $|\alpha_i|^2$ representing its relative weight. \\
\midrule
Collapse & The convergence of the superposed state to a dominant explanation or synthesized hybrid when coherence exceeds a threshold. \\
\midrule
Projection & The semantic alignment between an observation and a hypothesis, computed as cosine similarity in the embedding space. \\
\midrule
Entanglement & In our context (distinct from physics), the semantic interdependence between hypotheses where their meanings and explanatory power shift based on co-activation. \\
\midrule
Mix Operator & The mathematical function that combines high-amplitude hypotheses into hybrid explanations through weighted semantic blending. \\
\midrule
Coherence & A measure of how concentrated the explanatory state is, typically $C(\Psi) = \max_i |\alpha_i|^2$. \\
\midrule
Epistemic vs.\ Ontological & Our superposition is epistemic (about our knowledge/uncertainty), not ontological (about reality creating multiple worlds). \\
\midrule
$\kappa_{ij}$ & Interference coefficient encoding domain-specific interaction strength between hypotheses $i$ and $j$. \\
\bottomrule
\end{tabular}
\end{table}

\section{Formal Framework and Derivations}
\label{app:formal}

\subsection{State, Projection, and Interference}
\label{app:state}

\paragraph{State Representation} 
\vspace{-9pt}
\begin{equation}
\ket{\Psi^{(t)}} = \sum_{i=1}^n \alpha_i^{(t)} \ket{H_i}, \qquad \sum_{i=1}^n |\alpha_i^{(t)}|^2 = 1
\label{eqA:state}
\end{equation}

\paragraph{Semantic Embedding}
\vspace{-9pt}
\begin{equation}
\mathbf{h}_i, \mathbf{o}_j \in \mathbb{R}^d, \qquad
\mathbf{h}_i = \text{SBERT}(\text{text}(H_i)), \;\; \mathbf{o}_j = \text{SBERT}(\text{text}(O_j))
\label{eqA:embed}
\end{equation}

\paragraph{Projection Operator}
\vspace{-9pt}
\begin{equation}
\langle H_i | O_j \rangle = \cos(\mathbf{h}_i, \mathbf{o}_j) = \frac{\mathbf{h}_i \cdot \mathbf{o}_j}{\|\mathbf{h}_i\| \|\mathbf{o}_j\|}
\label{eqA:proj}
\end{equation}

\paragraph{Interference Matrix}
\vspace{-9pt}
\begin{equation}
I_{ij} = \kappa_{ij} \cdot \text{sim}(H_i, H_j), \qquad \kappa_{ij} \in [-1,1]
\label{eqA:Iij}
\end{equation}

\subsection{Amplitude Dynamics and Coherence}
\label{app:dynamics}

\paragraph{Update Rule and Normalization}
\vspace{-6pt}
\begin{equation}
\tilde{\alpha}_i^{(t+1)} = \alpha_i^{(t)} + \eta \left[ e_i^{(t)} + \sum_{k \neq i} I_{ik} \alpha_k^{(t)} \right],
\qquad
\alpha^{(t+1)}=\frac{\tilde{\alpha}^{(t+1)}}{\|\tilde{\alpha}^{(t+1)}\|_2}
\label{eqA:update}
\end{equation}

\paragraph{Coherence and Collapse}
\vspace{-6pt}
\begin{equation}
C(\Psi) = \max_{i} |\alpha_i|^2, \qquad \text{collapse if } C(\Psi) > \tau \text{ or when a decision is required.}
\label{eqA:collapse}
\end{equation}

\subsection{Mix Operator and Synthesis}

\paragraph{Mix Operator}
\vspace{-6pt}
\begin{equation}
\text{mix}(H_i, H_j) = \lambda_{ij} \mathbf{h}_i + (1 - \lambda_{ij}) \mathbf{h}_j, \qquad
\lambda_{ij} = \frac{|\alpha_i|^2}{|\alpha_i|^2 + |\alpha_j|^2}
\label{eqA:mix}
\end{equation}

For natural language articulation, we utilize template-based synthesis for simple merges and LLM-based critique to refine complex hybrids.

\subsection{Implementation Sketch and Complexity}
\label{app:impl}

\paragraph{Reference Sketch (pseudo-code).}
\begin{verbatim}
Input: alpha[1..n], I[1..n,1..n], evidence e[1..n], eta in (0,1]
for i in 1..n:
    alpha_tilde[i] = alpha[i] + eta * ( e[i] + sum_{k != i} I[i,k] * 
    alpha[k] )
Normalize: alpha = alpha_tilde / ||alpha_tilde||_2
Collapse if max_i |alpha[i]|^2 > tau; else continue
\end{verbatim}

\paragraph{Estimating $I_{ij}$.}
Initialize from cosine similarity; set $\kappa_{ij}$ via domain priors, weak supervision (optimize ECS/HQI on solved cases), or expert elicitation.

\paragraph{Complexity.}
Per step $O(n^2)$ for interference + $O(n)$ for projection aggregation; with $m$ observations, $O(n^2 m)$.
\section{Computational Sketch and Implementation Notes }
\label{app:sketch}
\subsection{State, Evidence, Interference}

Let $\mathcal{H}=\{H_i\}_{i=1}^n$ and $\mathcal{O}=\{O_j\}_{j=1}^m$. Maintain
\[
\ket{\Psi^{(t)}}=\sum_{i=1}^n \alpha_i^{(t)} \ket{H_i}, \qquad \sum_i |\alpha_i^{(t)}|^2=1.
\]
Embed 
 hypotheses and observations with a sentence encoder (e.g., Sentence-BERT \cite{reimers2019sentencebert}) to obtain vectors $\mathbf{h}_i,\mathbf{o}_j\in\mathbb{R}^d$. Use cosine similarity as a proxy for $\langle H_i\mid O_j\rangle$.

Evidence projection for a new batch is as follows $\mathcal{O}^{(t)}$:
\[
e_i^{(t)} \;=\; \max_{O_j\in \mathcal{O}^{(t)}} \frac{\mathbf{h}_i\cdot \mathbf{o}_j}{\|\mathbf{h}_i\|\,\|\mathbf{o}_j\|}.
\]

Interference matrix $I\in\mathbb{R}^{n\times n}$ models hypothesis interaction. A simple, effective choice is the following:
\[
I_{ii}=1,\qquad I_{ij}=\mathrm{cos\_sim}(\mathbf{h}_i,\mathbf{h}_j)\cdot \kappa_{ij},\;\; i\neq j,
\]
with $\kappa_{ij}\in[-1,1]$ permitting constructive ($>0$) or destructive ($<0$) coupling. In data-poor settings set $\kappa_{ij}=1$ (pure similarity). When expert priors are available, hand-tune signs to encode known exclusivities or complementarities.

\subsection{Amplitude Update and Collapse}

With learning rate $\eta\in(0,1)$
\[
\tilde{\alpha}_i^{(t+1)} \;=\; \alpha_i^{(t)} + \eta\Big(e_i^{(t)} + \sum_{k\neq i} I_{ik}\,\alpha_k^{(t)}\Big),\qquad
\alpha_i^{(t+1)} \;=\; \frac{\tilde{\alpha}_i^{(t+1)}}{\sqrt{\sum_\ell |\tilde{\alpha}_\ell^{(t+1)}|^2}}.
\]
Define a coherence functional that grows when one (or a consistent subset) dominates:
\[
\mathcal{C}^{(t)} \;=\; \max_i |\alpha_i^{(t)}|^2
\quad\text{or}\quad
\mathcal{C}^{(t)} \;=\; \sum_i |\alpha_i^{(t)}|^4.
\]
Collapse when $\mathcal{C}^{(t)}\ge \tau$ or when a decision deadline arrives. If several $|\alpha_i|$ remain high and $I_{ij}>0$, return a \emph{hybrid synthesis} $H_{i\wedge j}$.

\subsection{Synthesis operator (hybrids)}

Given two high-mass hypotheses $H_i,H_j$, form a content-level blend
\[
\mathrm{mix}(\mathbf{h}_i,\mathbf{h}_j)=\lambda\,\mathbf{h}_i+(1-\lambda)\,\mathbf{h}_j,
\quad \lambda=\frac{|\alpha_i|^2}{|\alpha_i|^2+|\alpha_j|^2},
\]
and instantiate a natural-language synthesis using a controlled LLM prompt that lists (i)~active hypotheses and amplitudes, (ii) salient observations and projections, (iii) entries of $I_{ij}$ that justify merger or exclusion. This yields faithful, auditable text (see \cite{pareschi2023medical} for a clinical instance and \cite{ghisellini2025entangled} for a strategic instance).%

\subsection{Implementation Note}

A lightweight pipeline suffices the following:
\begin{enumerate}\itemsep2pt
\item Sentence-BERT embeddings for $\mathbf{h}_i,\mathbf{o}_j$ (384–768d).
\item Evidence scores $e_i$ via cosine similarity.
\item $I_{ij}$ from similarity; optionally sign $\kappa_{ij}$ with expert priors (exclusivity vs.\ complementarity).
\item Iterate the update; monitor $\mathcal{C}^{(t)}$; stop at $\tau$ or deadline.
\item If hybrid, call the synthesis operator with $(\alpha,I,e)$ as structured context.
\end{enumerate}

.


\end{document}